\documentclass[journal]{IEEEtran}

\usepackage{framed,multirow}
\usepackage{amssymb}
\usepackage{latexsym}
\usepackage{graphicx}
\usepackage{booktabs}
\usepackage{amssymb}
\usepackage{pifont}
\usepackage{graphicx}
\usepackage{amsmath}
\usepackage{amssymb}
\usepackage{booktabs}
\usepackage{multirow}
\usepackage{multicol}
\usepackage{xcolor}
\usepackage{tabularx}
\usepackage{float}
\usepackage{algorithmic}
\usepackage{color, colortbl}
\usepackage{multirow}
\usepackage{multicol}
\usepackage{booktabs}
\usepackage{xcolor}
\usepackage{url}
\usepackage[normalem]{ulem}
\usepackage{url}
\usepackage[square, numbers]{natbib}
\usepackage{xcolor}
\usepackage{footnote}
\usepackage{etoolbox}

\newcommand{\etal}{\textit{et al.}}

\setcitestyle{square,aysep={,}}

\definecolor{newcolor}{rgb}{.8,.349,.1}
\newcommand{\cmark}{\ding{51}}
\newcommand{\xmark}{\ding{55}}
\setcitestyle{square,aysep={,}}

\ifCLASSINFOpdf
\else
\fi
\begin{document}
\title{FSBI: Deepfakes Detection with \\Frequency Enhanced Self-Blended Images}

\author{\IEEEauthorblockN{Ahmed Abul Hasanaath, Hamzah Luqman, Raed Katib, Saeed Anwar} \\
\IEEEauthorblockA{Information and Computer Science Department, King Fahd University of Petroleum and Minerals\\ SDAIA-KFUPM Joint Research Center for Artificial Intelligence, KFUPM, Saudi Arabia. \\
Email: {{g202302610, hluqman, g202212980, saeed.anwar}@kfupm.edu.sa}}
 
}

\maketitle

\begin{abstract}
Advances in deepfake research have led to the creation of almost perfect manipulations undetectable by human eyes and some deepfakes detection tools. Recently, several techniques have been proposed to differentiate deepfakes from realistic images and videos. 
This paper introduces a Frequency Enhanced Self-Blended Images (FSBI) approach for deepfakes detection. This proposed approach utilizes Discrete Wavelet Transforms (DWT) to extract discriminative features from the self-blended images (SBI) to be used for training a convolutional network architecture model. The SBIs blend the image with itself by introducing several forgery artifacts in a copy of the image before blending it. This prevents the classifier from overfitting specific artifacts by learning more generic representations. These blended images are then fed into the frequency features extractor to detect artifacts that can not be detected easily in the time domain.    
The proposed approach has been evaluated on FF++ and Celeb-DF datasets and the obtained results outperformed the state-of-the-art techniques with the cross-dataset evaluation protocol. The code is available at \url{https://github.com/gufranSabri/FSBI}.
\end{abstract}


%
\IEEEpeerreviewmaketitle

\section{Introduction}
\label{sec:introduction}
The rise of deepfake technology has introduced a new era of digital manipulation that poses severe threats to the authenticity and integrity of multimedia content. Deepfakes, which are artificially generated media that convincingly mimic real individuals or events, have the potential to deceive, manipulate, and spread misinformation on an unprecedented scale. From fake news and malicious content creation to privacy invasion and identity theft, the implications of deepfakes are multifaceted and alarming.

Several techniques have been proposed in the literature for generating fake images that can be synthesized entirely or partially. Artificial neural networks~\cite{rana_deepfake_2022}, auto-encoding variation~\cite{kingma_auto-encoding_2022,kingma_introduction_2019}, and Generative Adversarial Networks (GAN)~\cite{goodfellow_generative_2014} are playing a significant role in the development and creation of modified and fake images. More realistic fake images have been created using GAN variants~\cite{tolosana_deepfakes_2020}, such as Progressive GAN~\cite{karras2017progressive}, StyleGAN~\cite{karras2019style}, and SNGAN~\cite{miyato2018spectral}.
Other techniques have been proposed for real-time face manipulations, such as Face Swap~\cite{deepfakes_deepfakes_faceswap_2023}, Neural Textures~\cite{thies_deferred_2019}, and Face2Face~\cite{thies_face2face_2020}.

\begin{figure}[t]
    \centering
    \includegraphics[width=0.7\columnwidth]{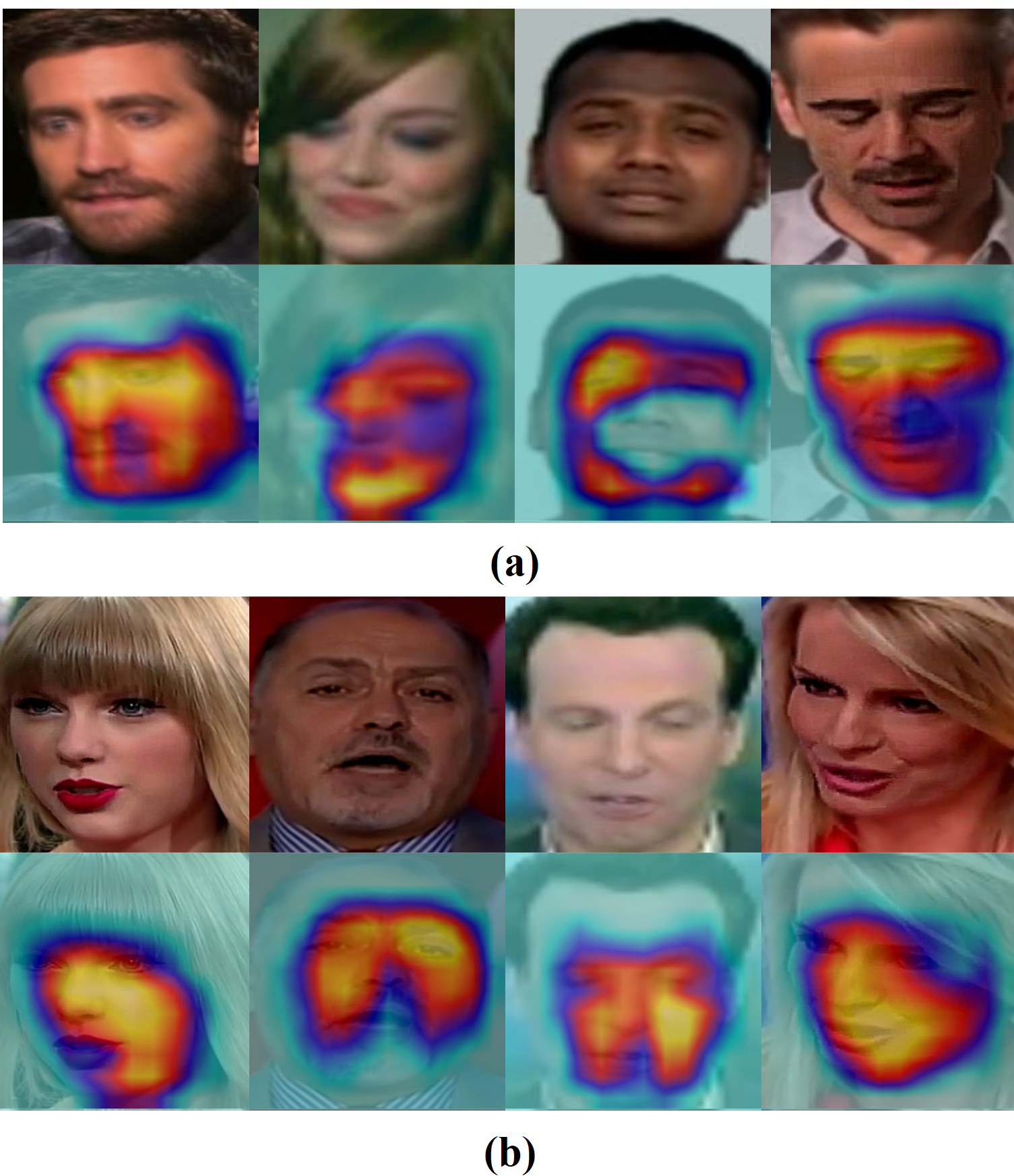}
    \caption{Examples of the artifact regions detected by the proposed FSBI approach on samples from (a) Celeb-DF and (b) FF++ datasets. }
    \label{fig:cam}
\end{figure}

Detecting these deceptive digital creations is a complex and pressing challenge in today's information-driven society. Traditional methods for identifying deepfakes have become increasingly ineffective as the sophistication of generative models and manipulation techniques continues to evolve. In this context, the quest for innovative and robust deepfake detection mechanisms becomes imperative to safeguard the credibility of digital media and maintain public trust.

In the dynamic field of deepfake detection, diverse techniques have emerged, combining traditional methods with cutting-edge deep learning approaches. Traditional techniques depend on examining facial expressions, eye movements, and lip sync for inconsistencies. In addition, visual facial artifacts, such as eyes lacking realistic reflections and vibrance or teeth appearing as white blobs, have been exploited for detection~\cite{li2018exposing, sahla2021detection, matern2019exploiting, yang2019exposing}. These artifacts can be captured easily in high-resolution images; however, downsampling the image size to meet computational constraints usually results in difficulties in capturing these artifacts and sometimes introduces other resolution inconsistencies. Therefore, some researchers used preprocessing techniques to address this issue and enhance model generalization by capturing a range of visual artifacts~\cite{cozzolino2018forensictransfer, xuan2019generalization}.
However, utilizing traditional techniques for deepfake detection faces challenges with advanced deepfake generation algorithms.

Recently, researchers leveraged deep learning techniques to scrutinize patterns in facial features, lighting, and background elements to detect deepfakes. Zhao~\etal proposed a multi-attentional deepfake detection network incorporating regional independence loss and attention-guided data augmentation~\cite{zhao2021multi}. Sun~\etal~\cite{sun2021improving}~introduced LRNet as a novel deepfake detection approach, while Zhao~\etal~\cite{zhao2023proactive}~improved accuracy by integrating anti-Deepfake labels into facial identity features. 

Shiohara~\etal~\cite{shiohara2022detecting} proposed deepfakes data synthesized using a self-blended images (SBIs) approach. For each original image, pseudo-source and target images are generated by applying several transformations to the original image. These images are blended to create the SBIs used for deepfakes detection. This paper proposes an enhancement to the approach proposed in~\etal~\cite{shiohara2022detecting}. We propose a Frequency Enhanced SBIs (FSBIs) approach for deepfakes detection. This approach extends the SBI method by extracting features from the SBIs in the frequency domain that help develop a more generic and robust detector. We fed each channel of blended images resulting from SBI into Discrete Wavelet Transforms (DWT). The resulting features are fused and fed into a Convolutional Neural Network (CNN) model to learn the artifacts in the deepfake images. We used within-dataset and cross-dataset evaluation protocols to evaluate the proposed approach and new state-of-the-art results were obtained from the FF++ and Celeb-DF datasets. The proposed approach showed a good capability for detecting the artifacts on the deepfake images, as shown in Figure~\ref{fig:cam}. We claim the following contributions in this paper:
\begin{itemize}
\item Proposing an innovative approach for deepfakes detection 
\item Discussing the efficiency of frequency-based transforms for detecting artifacts on deepfake images 
\item Evaluating the generalization of the proposed approach by performing cross-dataset evaluation
\item Reporting a new state-of-the-art performance on two benchmarking datasets
\end{itemize}

The rest of this paper is organized as follows: Section II surveys the literature of deepdake detection and Section \ref{sec:approach} presents the framework of the proposed approach. Section \ref{sec:experiments} presents experimental work and the results of our work along with comparisons and discussions. Finally, section \ref{sec:conclusion} concludes the paper.

\section{Related Works} \label{sec:litreview}
Several techniques have been proposed in the literature for deepfakes detection. These techniques vary from simple computer vision algorithms to complex deep-learning neural networks. Therefore, we classify these techniques into traditional and deep learning approaches. 

\vspace{2mm}\noindent\textbf{Traditional Techniques. }
These techniques depend mainly on classic computer vision operations to detect artifacts. Although the traditional methods may sound old school, they are recently employed, such as Kang~\etal~\cite{kang_detection_2022}, which used residual noise, wrapping artifacts, and blurring effects to detect deepfakes. Jeong~\etal~\cite{jeong_bihpf_2022} employed bilateral high-pass filters to detect amplification effects on fake images. Another deepfake detector was proposed by Liu~\etal~\cite{liu_ti2net_2023} utilizing the temporal identity inconsistency between human faces among video frames. 
Although these techniques showed good accuracy for detecting deepfakes, the significant advance in the deepfake generation motivated researchers to look for alternative techniques using deep learning~\cite{verdoliva_media_2020}. 

\vspace{2mm}\noindent\textbf{Deep Learning Techniques. } 
Most of the recent techniques depend on neural networks for deepfake detection \cite{nguyen2022deep}. The majority of these techniques~\cite{ilhan_improved_2022, khormali_dfdt_2022, luo_generalizing_2021,xu_learning_2023, zhao_learning_2021, nadimpalli_improving_2022} utilized CNNs for learning and classifying images into real and fake classes. The majority of these CNN models did not undergo end-to-end training, instead relying on pre-trained models for feature extraction. The EfficientNet models are the most dominant pre-trained models used in most of the CNN-based deepfake detection techniques~\cite{jevnisek_aggregating_2022, coccomini_cross-forgery_2022, wang_deep_2022}. To boost the accuracy of the deepfake detection techniques, several researchers utilized CNN with the transformer model~\cite{jevnisek_aggregating_2022, coccomini_combining_2022, coccomini_cross-forgery_2022, wang_deep_2022, wodajo_deepfake_2021, shiohara2022detecting}.
Wodajo~\etal~\cite{wodajo_deepfake_2021} used CNN to extract spatial features and fed them into a vision transformer (ViT) model that uses an attention mechanism to categorize them into real or fake images. Similarly, CNN was combined with the ViT by other researchers~\cite{coccomini_combining_2022, coccomini_cross-forgery_2022, wang_deep_2022, heo_deepfake_2021} and good results have been reported.

Chen~\etal~\cite{chen_local_2021} proposed a face forgery detection method by learning the local features and detecting inconsistencies. To address the generalization issue of the current deepfakes detection methods, Chen~\etal~\cite{chen_self-supervised_2022} employed a self-supervised learning-based method. The model employs a pool of forgery configurations, along with synthesized augmented forgeries, to detect forgeries in images and videos. Lee~\etal~\cite{lee_tar_2021} presented a digital forensics tool for detecting different types of deepfakes. The proposed tool learns the features using an autoencoder-based model.

The generalization of the deepfake detection techniques is essential for developing real-time systems. Therefore, researchers usually perform cross-dataset evaluation with the proposed deepfake detection approaches~\cite{coccomini_cross-forgery_2022, tolosana_deepfakes_2020, khormali_dfdt_2022, ge_explaining_2021, luo_generalizing_2021,xu_learning_2023, zhao_learning_2021}. In this setting, the technique is evaluated on a dataset different than the dataset used for model training. Coccomini~\etal~\cite{coccomini_cross-forgery_2022} performed cross-forgery analysis using ViT to detect deepfakes in various datasets. Luo~\etal ~\cite{luo_generalizing_2021} used high-frequency features and CNNs to generalize the detection methods. The MCX-API~\cite{xu_learning_2023} method used a pairwise learning method to improve the model's generalization. Deepfake Disrupter~\cite{wang_deepfake_2022} was used with the help of human input and deepfake detection tools to detect deepfakes. However, most of the proposed techniques cannot detect deepfakes that are not present in the training model~\cite{verdoliva_media_2020}.

\section{Frequency Enhanced Self-Blended Images (FSBI)}
\label{sec:approach}
\begin{figure*}[tbp]
    \centering
    \includegraphics[width=\linewidth]{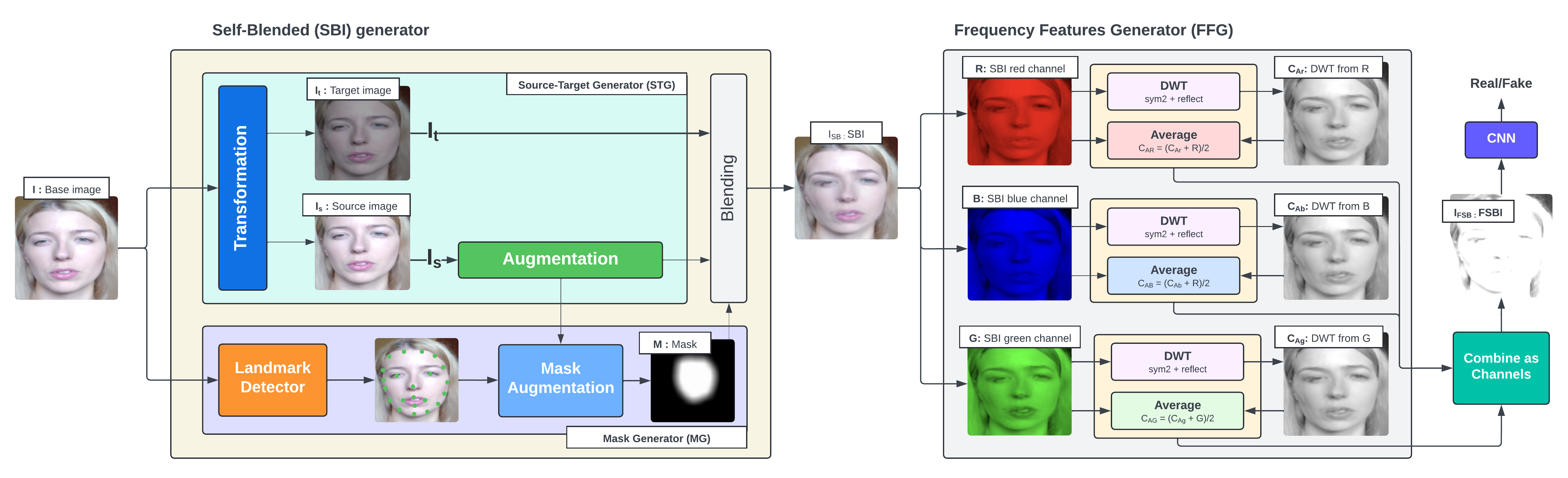}
    \caption{\textbf{The framework of the FSBI model.} The process begins with creating an SBI. The SBI is then decomposed into R, G, and B channels by the FFG. DWTs are computed for each channel individually, and the approximate coefficients are obtained. The approximate coefficients of each channel are then combined with the original channel through a simple averaging operation. These channels are stacked channel-wise final FSBI image. The resulting FSBIs are used to train a CNN classifier to recognize real and fake images.  }
    \label{fig:approach}
\end{figure*}

We propose an FSBI approach for deepfakes detection and its framework is shown in Figure~\ref{fig:approach}. The proposed approach comprises three modules: an SBI generator, a Frequency Features Generator (FFG), and a CNN classifier. The original image is fed into the SBI generator component to create a blended image with some artifacts. The FFG applied a series of DWT wavelets to the resulting blended image to extract some features in the frequency domain. The features are then fused and fed into a CNN classifier trained with an EfficientNet-B5 pre-trained model to learn and classify images as real or fake.

\subsection{Self-Blended Image (SBI)}
SBIs are images that are generated by blending the image with itself. Shiohara~\etal proposed the idea of SBI~\cite{shiohara2022detecting}, which depends on creating deepfakes by blending the face with its mask, generated by applying several transformations. The SBI method follows three primary stages to generate deepfakes from a given original image. First, a source-target generator (STG) component takes an image as an input and generates pseudo source ($I_{s}$) and target ($I_{t}$) images. These images are generated by performing random image transformations to the original image. These transformations involve manipulating the values of RGB and HSV channels and changing the brightness and contrast of the image. Additional augmentation operations involve downsampling and translation are applied to the source image to create landmark mismatches.

In the second stage, the SBI uses a mask generator (MG) to provide a gray-scale mask image that blends both the source and target images. The MG generates a mask by detecting the landmarks of the facial region and then calculating the convex hull from the detected landmarks. For this process, the SBI method adopts the FaceXray BI~\cite{li2019face} to perform landmark detection followed by elastic deformation~\cite{zhao_learning_2021}. The masks are smoothed twice using two Gaussian filters while being eroded after the first filter is applied.

Finally, the SBI blends the source and target images with the generated mask to create the SBI. The blended image is created by eroding the source image with the mask generated by the MG module. The resulting eroded image is added to the resulting image from eroding the target image with the inverse of the mask. These steps can represented as follows:
 
\begin{equation}
\label{eq:isb}
\mathrm{I_{SBI}} = \mathrm{I_{s}} \odot \mathrm{M} + \mathrm{I_{t}} \odot (1 - \mathrm{M})
\end{equation}
where $I_{s}$ is the source image, $I_{t}$ is the target image, and $M$ is the generated mask.

\subsection{Frequency Features Generator (FFG)} \label{sec:FSBI}

The blended image resulting from the SBI module is fed into the FFG module. This module leverages DWT to extract more discriminative features from the blended images that can help in detecting the artifacts on the deepfake images. 
DWT is a signal and image processing technique that decomposes a signal into different frequency components. It operates by iteratively applying filters to the signal and dividing it into approximation and detail coefficients at various resolution levels. This process provides a multi-resolution analysis that captures both global and local features within the image. DWT families, such as Haar, Daubechies, Symlet, Biorthogonal, and Coiflet, offer different properties suited for various applications. Haar is computationally efficient but does not perform well for complex structures. Daubechies, with varying orders, are effective for fine details and compression, whereas Symlet balances smoothness and oscillations and it is suitable for signals with both smooth and abrupt changes. Biorthogonal offers flexibility with separate sets for decomposition and reconstruction, while Coiflet, with improved regularity, is fitting for applications requiring smoother wavelets. The choice among these families depends on the specific characteristics of the deepfake images and the requirements of the detection task. Therefore, we evaluated Haar, Symlet, and Biorthogonal types for deepfakes detection and the best results were obtained using Symlet wavelet type. Additionally, 
several signal extension modes can be
applied to reduce the artifacts resulting from signal extrapolation when computing the DWT. These extension modes
include Symmetric, Reflect, and Antireflect. We explored all of these modes in this work and the highest score was obtained using the Reflect mode.

The RGB channels of the SBI are decomposed into three channels (R, G, and B) and the DWT coefficient is computed for each channel. The computed approximate coefficient of each channel has been resized to the size of the original images and averaged with the corresponding SBI channel.   
The resulting coefficients are concatenated depthwise to create a three-channel DWT image, which we refer to as FSBI. This image highlights all possible artifacts introduced in the deepfake during its generation.   

\subsection{CNN Model}
The last component of the proposed FSBI approach consists of a CNN model. The model is trained to detect artifacts on the FSBIs and classify images as real or fake. The model is trained on the FSBIs that encode the SBI with the frequency features extracted from each channel of the SBI. We fine-tuned an EfficientNet-B5~\cite{tan2019efficientnet} pre-trained on ImageNet for this task. EfficientNet-B5, a member of the EfficientNet family, is characterized by compound scaling, where width, depth, and resolution are uniformly scaled to balance model capacity and computational cost effectively. Its design prioritizes SOTA accuracy on image classification tasks, such as ImageNet, with fewer parameters compared to other models of similar performance.

\section{ Experiments} \label{sec:experiments}

\subsection{Experimental Setup} \label{sec:implementation}

\vspace{2mm}\noindent\textbf{Datasets.}  
We employed two prominent datasets to evaluate the proposed approach: FF++~\cite{rossler2019faceforensics++} and Celeb-DF~\cite{li2020celeb}. The FF++ dataset consists of 1,000 videos with around 500,000 frames. This dataset has fake images created using deepfake (DF)~\cite{githubGitHubDeepfakesfaceswap}, face swap (FS)~\cite{thies2016face2face}, neural textures (NT)~\cite{thies2019deferred}, and Face2Face (F2F)~\cite{githubGitHubMarekKowalskiFaceSwap} methods. From the 1,000 real videos, FF++ constructed over 1.8 million images derived from 4,000 fake videos. Due to the large size of the FF++ dataset, we used the compressed c23 version of the FF++ dataset (FF++(LQ)) in this work. We also used the Celeb-DF (version 2) dataset, which consists of large-scale deepfake videos generated using specialized synthesis methods. The dataset comprises 5,639 high-quality deepfake videos featuring celebrities, equivalent to 2 million frames.

\vspace{2mm}\noindent\textbf{Implementation Details.} The model was trained and evaluated on an Ubuntu 22.04 LTS machine equipped with an NVIDIA RTX A5000 GPU with 48 GB of memory. A PyTorch framework was used to implement the proposed approach. The proposed model has been trained with an early stopping technique to avoid overfitting. However, to have a valid comparison with the baseline model, the SBI method~\cite{shiohara2022detecting}, we employed the same number of epochs (100 epochs) to train the SBI method.

\vspace{2mm}\noindent\textbf{Training.} For the CNN model, we fine-tuned the EfficientNet-B5 model that was pre-trained on ImageNet. We employed the Sharpness-Aware Minimization (SAM) optimizer to optimize the training process. We set the momentum for the optimizer to 0.9 and the learning rate to 0.001. The learning rate linearly decays after 75 epochs. We train the model on a batch size of 16. The size of the image is set to 380$\times$380.

\vspace{2mm}\noindent\textbf{Inference.} In the inference stage, a set consisting of 32 frames is extracted from each video. If the frame contains two or more faces, the classifier assesses each face, and the confidence associated with the highest fakeness is considered the predicted confidence for that frame. After obtaining predictions for all frames, we calculate the average to determine the video's overall prediction. To ensure a fair comparison, we evaluate all videos in the test sets, setting confidence to 0.5 for videos when no faces are detected in any frames.

\subsection{Results} \label{sec:results}
This section presents the evaluation results of the FSBI approach proposed for deepfake detection. Two settings have been followed for evaluating our approach: within-dataset and cross-dataset evaluations. We used the Area Under the Curve (AUC) metric in both settings. 

\vspace{2mm}\noindent\textbf{Within-dataset Evaluation.}
In this evaluation, the performance of the FSBI is evaluated on the same dataset on which it was trained. Table~\ref{tab:crossDatasets} presents the obtained results with the FF++ and Celeb-DF datasets. The deepfake images of the FF++ dataset are available in four formats based on the used deepfake generation technique (DF, FS, F2F, and NT), as discussed in Section~\ref{sec:implementation}. We evaluated the performance of the FSBI on all these versions of FF++. Our model reported a high AUC with both datasets, as shown in the top part of Table~\ref{tab:crossDatasets}. For FF++, the FSBI approach can recognize deepfakes generated using almost all manipulation methods. 

A high AUC is obtained with the FF++ dataset when the deepfake images are generated using the DF method. In contrast, a low AUC is obtained when the NT manipulation technique is used, and this can be attributed to the fact that the NT method introduces a novel paradigm for image synthesis. NTs are learned feature maps that store high-dimensional information on top of 3D mesh proxies, which significantly differs from traditional textures used in other deepfake generation techniques. This unique representation likely presents challenges for the deepfake detection model, causing it to perform less effectively. In addition, the compressed videos of the FF++ dataset (FF++(LQ)) used in this work are of low resolution. Detecting deepfakes in low-resolution images is more challenging than in raw, high-resolution images. For the Celeb-DF dataset, a high AUC is obtained when the approach is trained and evaluated using the samples from the same dataset.

\vspace{2mm}\noindent\textbf{Cross-dataset Evaluation.}
A crucial aspect of assessing the generalization capabilities of our approach is to evaluate its performance across different data distributions and manipulation techniques. This type of evaluation is called a cross-dataset evaluation and serves as a robust test to gauge the effectiveness and adaptability of any deepfake detection approach in real-world scenarios. In this evaluation, the dataset used to test the model differs from the dataset used for training. We evaluated our FSBI using this evaluation setting, and the results are shown in Table \ref{tab:crossDatasets}. Notably, when the model is trained on the FF++ dataset and tested on the Celeb-DF dataset, it outperforms related works, achieving an impressive AUC of 95.49\%. Furthermore, training the model on the Celeb-DF dataset yields high results when evaluated on the FF++ dataset. We evaluated the model on the FF++ dataset with different manipulation techniques. The table shows that the FSBI achieved a high AUC on deepfakes generated using DF and FS methods. However, the AUC scores for the FSBI model with the F2F and NT methods are relatively lower. These results with NT manipulations align with those obtained within-dataset evaluation setting.

\begin{table}[tbp]
  \centering
  \caption{The AUC of the FSBI approach with within- and cross-dataset evaluations. }
  \label{tab:crossDatasets}
\resizebox{\columnwidth}{!}{
\begin{tabular}{lllccccc}
\toprule
\multicolumn{1}{c}{\multirow{2}{*}{\textbf{Evaluation}}} & \multicolumn{1}{c}{\multirow{2}{*}{\textbf{Train}}} & \multicolumn{1}{c}{\multirow{2}{*}{\textbf{Test}}} & \multicolumn{4}{c}{\textbf{Manipulation}}                                                                            & \multirow{2}{*}{\textbf{Average}} \\ \cmidrule{4-7}
\multicolumn{1}{c}{}                                     & \multicolumn{1}{c}{}                                & \multicolumn{1}{c}{}                               & \multicolumn{1}{c}{\textbf{DF}} & \multicolumn{1}{c}{\textbf{FS}} & \multicolumn{1}{c}{\textbf{F2F}} & \textbf{NT} &                                   \\ \midrule
\multirow{2}{*}{\textbf{Within-dataset}}                            & FF++                                                 & FF++                                                & \multicolumn{1}{c}{99.45}       & \multicolumn{1}{c}{97.09}       & \multicolumn{1}{c}{95.37}        & 88.59       & 95.13                             \\   
                                                           & Celeb-DF                                             & Celeb-DF                                            & \multicolumn{1}{c}{}            & \multicolumn{1}{c}{}            & \multicolumn{1}{c}{}             &             & 95.40                             \\ \midrule 
\multirow{2}{*}{\textbf{Cross-dataset}}                            & Celeb-DF                                             & FF++                                                & \multicolumn{1}{c}{96.76}       & \multicolumn{1}{c}{95.43}       & \multicolumn{1}{c}{87.62}        & 78.07       & 89.47                             \\   
                                                           & FF++                                                 & Celeb-DF                                            & \multicolumn{1}{c}{}            & \multicolumn{1}{c}{}            & \multicolumn{1}{c}{}             &             & 95.49                             \\ \bottomrule
\end{tabular}
}
\end{table}

\begin{table}[h]
  \centering
  \caption{Our FSBI method demonstrates higher performance compared to other deepfake detection approaches. This is evident when we train and test on the same dataset (Within-DS: Within-dataset) and when trained on one dataset and evaluated on a different dataset (Cross-DS: Cross-dataset).}
  \label{tab:comp}
\footnotesize{\begin{tabular}{cllcc}
\toprule
\multirow{2}{*}{\textbf{Eval.}}    & \multirow{2}{*}{\textbf{Method}}          & \multirow{2}{*}{\textbf{Train}} & \multicolumn{2}{c}{\textbf{Test Dataset}}                                    \\ \cline{4-5} 
                               &                                           &                                & \textbf{FF++} & \textbf{Celeb-DF} \\ \toprule
\multirow{5}{*}{\rotatebox[origin=c]{90}{\textbf{Within-DS}}}    & Jevnisek et al. \cite{24jevnisek2022aggregating}          & FF++                           & 75.57         & -          \\  
                               & ViT \cite{coccomini2022combining}             & FF++                           & 95.10          & -          \\  
                               & MCX-API~\cite{51xu2023learning}           & FF++                       & \textbf{99.6 }        &                            \\  

                               & CNN+ViT \cite{50wodajo2021deepfake}               & FF++                           & 91.50          & -          \\  
                               & \textbf{FSBI (ours)  }                             & FF++                           & 95.13              & -          \\ \midrule
\multirow{9}{*}{\rotatebox[origin=c]{90}{\textbf{Cross-DS}}} & MCX-API~\cite{51xu2023learning}           & Multiple                       &           & 90.87                           \\  
                               & FRDM~\cite{35luo2021generalizing}         & Multiple                       & -             & 79.40                           \\  
                               & PCL + I2G~\cite{zhao2021learning}         & Multiple                       & -             & 90.03                           \\   
                               & LRL~\cite{zhao2021multi}                  & Multiple                       & -             & 78.26                           \\  
                               & LipForensics~\cite{19haliassos2021lips}   & FF++                           & -             & 82.40                           \\  
                               & FTCN~\cite{54zheng2021exploring}          & FF++                           & -             & 86.90                           \\  
                               & SBI (FF++(HQ))~\cite{shiohara2022detecting} & FF++                           & -             & 93.18                           \\  
                               & SBI (FF++(LQ))~\cite{shiohara2022detecting} & FF++                           & -             & 92.87                           \\  
                               & \textbf{FSBI (ours)}                      & FF++                           & -             & \textbf{95.49}                  \\ \bottomrule
\end{tabular}
}
\end{table}

\subsection{ Ablation Study} \label{sec:ablation}
We conducted ablation studies to verify the effectiveness of the proposed FSBI approach. We systematically varied key components and hyperparameters in our deepfake detection pipeline to understand their impact on performance while optimizing the accuracy of the proposed model. The components and hyperparameters are FFG, DWT type and mode, image size, and a pre-trained model. We conducted these ablation studies in the cross-dataset evaluation setting, using the FF++ dataset for training and the Celeb-DF for testing.

\begin{figure}[bph]
    \centering
    \includegraphics[width=0.9\columnwidth]{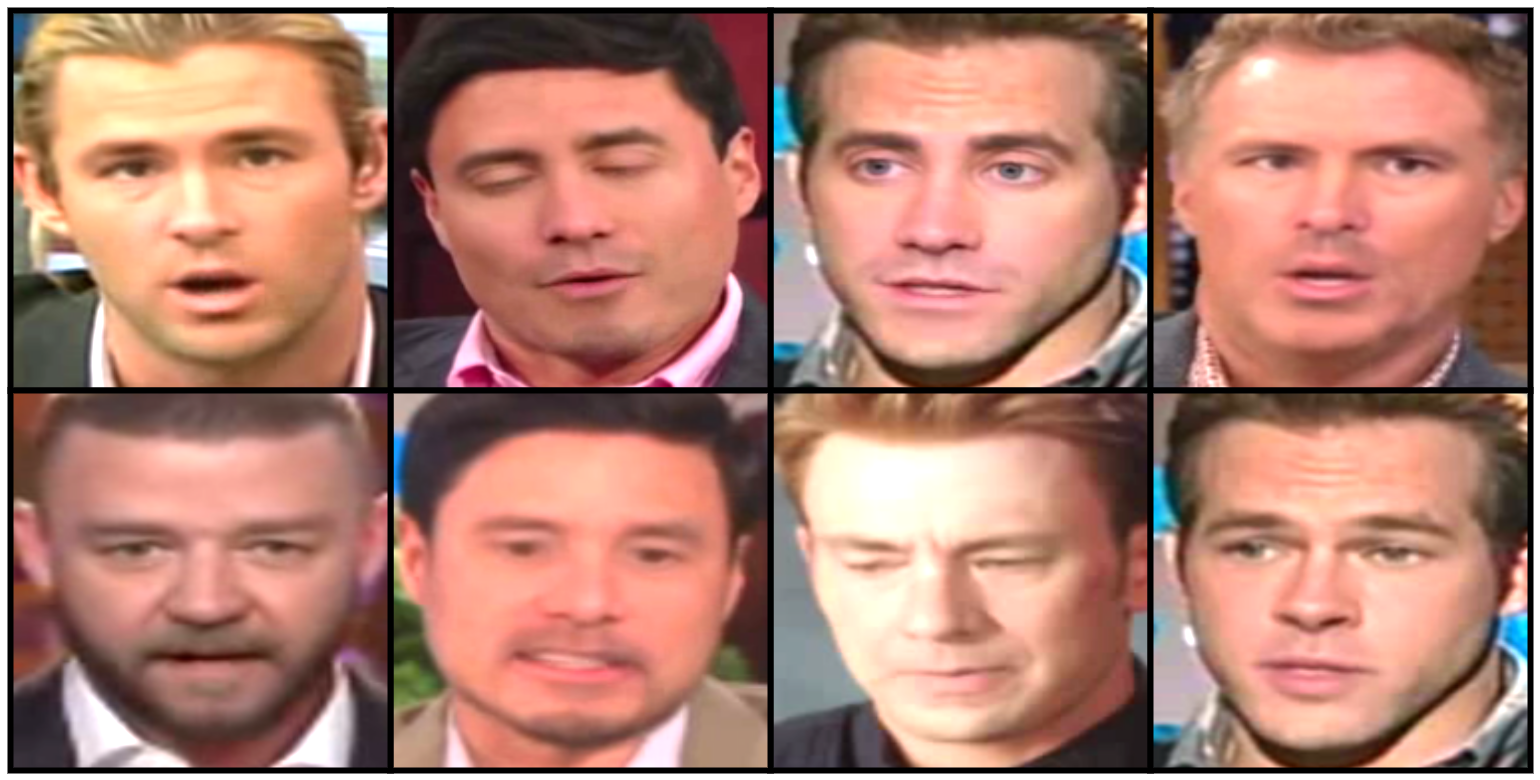}
    \caption{\textbf{Deepfake samples from Celeb-DF dataset.} The first row contains samples which were correctly detected as deepfakes by our FSBI approach whereas the SBI method failed. The second row contains samples that are challenging for both approaches. Both failed to detect them as deepfakes.}
    \label{fig:misclassified}
\end{figure}

\vspace{2mm}\noindent\textbf{FFG Component.} Our approach consists of three components: SBI, FFG, and CNN. We implemented a specific training procedure to evaluate the effectiveness of the first two components. We trained the model for 75 epochs on one component and fine-tuned it for 25 epochs on another component. We also trained and fine-tuned the model on both components, as shown in Table \ref{tab:ratio_exp}, where half of the images are generated using SBI while the other half contains images generated using FSBI. The objective was to identify the performance of the proposed approach with and without including the FFG module. The table shows that training and fine-tuning the model solely on FSBI resulted in the highest AUC score of 95.49\%. The results indicate an overall enhancement in model performance when trained on FSBI compared to SBI alone or when images generated by SBI are involved in either training or fine-tuning the model. This improvement can be attributed to the nature of the FFG, which employs the DWT to amplify the frequency of information in the SBI images. Training the model on images generated using FSBI makes it focus more on frequency artifacts present in deepfakes. Additionally, DWT introduces texture analysis-based features crucial for detecting deepfake artifacts resulting from sharp intensity changes during the generation process.

\begin{table}[tbp]
  \centering
  \caption{Results from experimenting the effectiveness of the framework components. Model was trained for 75 epochs and finetuned for 25 epochs.}
  \label{tab:ratio_exp}

  \begin{tabular}{cc|cc|c}
\toprule
\multicolumn{2}{c|}{\textbf{Training}}                      & \multicolumn{2}{c|}{\textbf{Fine-tuning}}                   &   \multirow{2}{*}{\textbf{AUC (\%)} }                     \\ \cmidrule{1-4}
\multicolumn{1}{c}{\textbf{SBI }} & \textbf{FSBI} & \multicolumn{1}{c}{\textbf{SBI }} & \textbf{FSBI} & \\ \midrule
\multicolumn{1}{c}{\cmark }    &     \textcolor{red}{ \xmark }          & \multicolumn{1}{c}{\cmark }    &       \textcolor{red}{ \xmark }        & 94.66                  \\  
\multicolumn{1}{c}{\cmark }    &       \textcolor{red}{ \xmark }        & \textcolor{red}{ \xmark }             & \cmark     &     94.78                   \\  
\textcolor{red}{ \xmark }            & \cmark     & \multicolumn{1}{c}{\cmark }    &      \textcolor{red}{ \xmark }         & 94.01                  \\ 
\textcolor{red}{ \xmark }        & \cmark     & \textcolor{red}{ \xmark }         & \cmark     & \textbf{95.49}         \\  
\multicolumn{1}{c}{\cmark }    & \cmark     & \multicolumn{1}{c}{\cmark }    &     \textcolor{red}{ \xmark }          & 94.21                  \\  
\multicolumn{1}{c}{\cmark }    & \cmark     & \textcolor{red}{ \xmark }             & \cmark     & 94.07                  \\  
\multicolumn{1}{c}{\cmark }    & \cmark     & \multicolumn{1}{c}{\cmark }    & \cmark     & 94.90                  \\ \bottomrule
\end{tabular}
\end{table}

\begin{table}[h]
  \centering
  \caption{The performance of the FSBI with different image sizes. }
  \label{tab:size_exp}
  \begin{tabular}{cc}
  \toprule
\textbf{Image Size (px)} & \textbf{AUC}   \\ \toprule
170 $\times$  170                    & 88.55          \\
230  $\times$  230                  & 92.55          \\
380  $\times$  380                  & \textbf{95.49} \\
512  $\times$  512                   & 91.59    \\  \bottomrule   
\end{tabular}
\end{table}

\vspace{2mm}\noindent\textbf{DWT Variables.} DWT plays an important role in the FSBI approach for representing and extracting features from SBI images in the frequency domain. Three main wavelet types of DWT are available (Biorthogonal, Symlet, and Hear), and the selection between them depends on the downstream task. In addition, three signal extension modes, Symmetric, Reflect, and Antireflect, can be applied to reduce the artifacts resulting from signal extrapolation when computing the DWT. To study the impact of the wavelet type and the signal extension modes on the performance of the FSBI, we conducted a cross-dataset evaluation and the obtained results are shown in Table~\ref{tab:wavelet_exp}. The obtained results reveal that the choice of DWT variables significantly influences the model's performance, with the best-performing configuration achieved using the Symlet wavelet and reflect mode ($M_5$ model), resulting in an AUC of 95.49\%. This emphasizes the importance of carefully selecting DWT parameters to optimize the model's ability to detect deepfake images. However, we can notice from these results that the obtained results with all DWT configurations still outperform the SOTA results on the Celeb-DF dataset. This supports our hypothesis of the efficiency of DWT for deepfake detection.

\vspace{2mm}\noindent\textbf{Image Size Selection.} The choice of image size is an essential factor in the performance of the FSBI approach. Therefore, we experimented with four image sizes: 170$\times$170, 230$\times$230, 380$\times$380, and 512$\times$512. As shown in Table \ref{tab:size_exp}, we observe notable differences in AUC scores when employing different image resolutions.
The results improve as the image size increases, which is expected since the artifacts will be more visible. 
We can also notice that the model seems to lose its ability to learn on a large image of size 512$\times$512. This can be attributed to information lost when applying an interpolation technique to enlarge the original image size from 256$\times$256 to 512$\times$512.

\vspace{2mm}\noindent\textbf{Choice of the Pretrained Model.} Although the CNN model utilized the pre-trained EfficientNet-B5 model in the FSBI framework, we evaluated the performance of the proposed approach using other network architectures. 
Our experiments incorporated EfficientNet-B4~\cite{tan2019efficientnet}, EfficientNet-B5, ResNet50~\cite{he2016deep}, and MobileNetV2~\cite{sandler2018mobilenetv2}. We also show the results of the standard SBI using these models. As shown in Table \ref{tab:model_exp}, EfficientNet-based models consistently outperformed ResNet50 and MobileNetV2 in various configurations. In addition, EfficientNet-B5 improved the performance of SBI and FSBI by more than 1\% compared with EfficientNet-B4. These results show that selecting the network architecture plays a vital role in improving the accuracy of the deepfake detector.

\begin{figure}[h]
    \centering
    \includegraphics[width=\columnwidth]{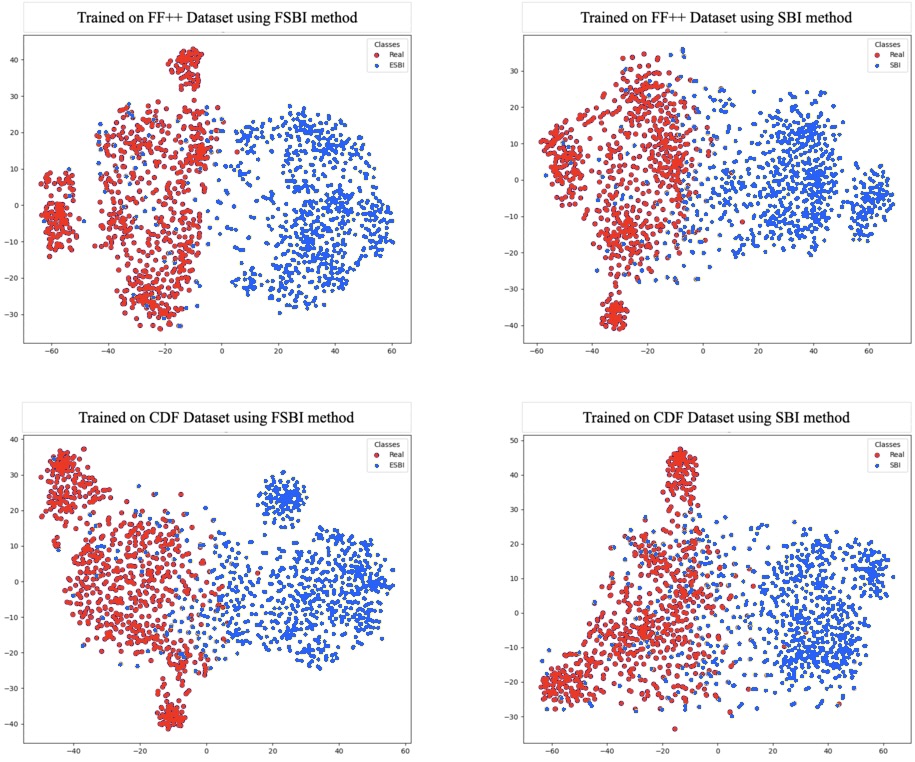}
    \caption{Feature space visualization of FSBI (first column) and SBI (second column) trained on (a) FF++ and (b) Celeb-DF datasets.}
    \label{fig:feature}
\end{figure}

\subsection{ Comparisons} \label{sec:comparison}
This section compares our FSBI approach with the SOTA deepfake detection techniques. In this comparison, two evaluation settings are followed: within-dataset and cross-datasets, as shown in Table \ref{tab:comp}. For within-dataset evaluation, FF++ was used to train and test models. The majority of deepfake detection techniques use this dataset for within-dataset evaluation. To our knowledge, the Celeb-DF dataset was used in the literature for cross-dataset evaluation.

We also performed a cross-dataset evaluation using Celeb-DF as a testing dataset. In contrast to other approaches in the literature that used multiple datasets for training \cite{51xu2023learning, 35luo2021generalizing, zhao2021learning, zhao2021multi}, our approach used only the FF++ dataset. To ensure a fair comparison with the SBI approach \cite{shiohara2022detecting}, we have reproduced their results using the compressed c23 version (FF++(LQ)) of the FF++ dataset. As shown in Table \ref{tab:comp}, our approach significantly outperformed all other techniques. Additionally, it outperformed approaches trained on multiple datasets. 

To visualize the qualitative difference between the proposed FSBI approach and the SBI method~\cite{shiohara2022detecting}, the t-SNE is applied to the feature vectors from the last layer of both models. Figure \ref{fig:feature} illustrates the feature space visualization of FSBI and SBI methods when trained on FF++ and Celeb-DF datasets. As shown in the feature visualization plot, when considering FF++, the FSBI method generates a distinct boundary between the clusters of the two classes. Conversely, the SBI method exhibits a less well-defined boundary between the two classes. In the context of Celeb-DF, the FSBI method demonstrates significantly reduced overlap between the two clusters compared to the SBI method. We also show in Figure \ref{fig:misclassified} some deepfake samples that were correctly detected by our approaches, whereas SBI failed to detect them. The figure also shows other samples, which are challenging for both methods.

\begin{table*}[th]
\centering
\caption{ Ablation study results with different DWT wavelet types and modes.}
\label{tab:wavelet_exp}
\begin{tabular}{c|ccc|ccc|c}
 \toprule
\multirow{2}{*}{\textbf{Model}}&\multicolumn{3}{c|}{\textbf{Wavelet}}                                                                                             & \multicolumn{3}{c|}{\textbf{Mode}}                                                                                                & \multirow{2}{*}{\textbf{AUC}} \\ \cline{2-7}
&\textbf{Biorthogonal}              & \textbf{Symlet}               & \textbf{Haar}               & \textbf{Symmetric}          & \textbf{Reflect}            & \textbf{Antireflect}        &                               \\  \midrule
$M_1$& \cmark  & \textcolor{red}{ \xmark }   & \textcolor{red}{ \xmark }   & \cmark  & \textcolor{red}{ \xmark }   & \textcolor{red}{ \xmark }   & 93.42                         \\ 
$M_2$&\cmark  & \textcolor{red}{ \xmark }   & \textcolor{red}{ \xmark }   & \textcolor{red}{ \xmark }   & \cmark  & \textcolor{red}{ \xmark }   & 93.97                         \\ 
$M_3$&\cmark  & \textcolor{red}{ \xmark }   & \textcolor{red}{ \xmark }   & \textcolor{red}{ \xmark }   & \textcolor{red}{ \xmark }   & \cmark  & 93.97                         \\ 
$M_4$&\textcolor{red}{ \xmark }   & \cmark  & \textcolor{red}{ \xmark }   & \cmark  & \textcolor{red}{ \xmark }   & \textcolor{red}{ \xmark }   & 94.87                         \\ 
$M_5$&\textcolor{red}{ \xmark }   & \cmark  & \textcolor{red}{ \xmark }   & \textcolor{red}{ \xmark }   & \cmark  & \textcolor{red}{ \xmark }   & \textbf{95.49}                \\ 
$M_6$&\textcolor{red}{ \xmark }   & \cmark  & \textcolor{red}{ \xmark }   & \textcolor{red}{ \xmark }   & \textcolor{red}{ \xmark }   & \cmark  & 95.21                         \\ 
$M_7$&\textcolor{red}{ \xmark }   & \textcolor{red}{ \xmark }   & \cmark  & \cmark  & \textcolor{red}{ \xmark }   & \textcolor{red}{ \xmark }   & 94.65    \\ 
$M_8$&\textcolor{red}{ \xmark }   & \textcolor{red}{ \xmark }   & \cmark  & \textcolor{red}{ \xmark }   & \cmark  & \textcolor{red}{ \xmark }   & 94.65    \\ 
$M_9$&\textcolor{red}{ \xmark }   & \textcolor{red}{ \xmark }   & \cmark  & \textcolor{red}{ \xmark }   & \textcolor{red}{ \xmark }   & \cmark  & 94.65    \\ \bottomrule
\end{tabular}
\end{table*}

\begin{table*}[h]
\centering
\caption{The performance of SBI and FSBI with different pre-trained models. Our proposed model consistently outperformed SBI for every each pre-trained models. }
  \label{tab:model_exp}
\begin{tabular}{c|cccc|c}
\toprule
\textbf{Models} & \textbf{ResNet-50} & \textbf{MobileNet-V2} & \textbf{EfficientNet-B4} & \textbf{EfficientNet-B5} & \textbf{AUC}            \\ \toprule
 SBI~\cite{shiohara2022detecting} & \textcolor{black}{\cmark}          &     \textcolor{red}{ \xmark }      &     \textcolor{red}{ \xmark }  &     \textcolor{red}{ \xmark }      & 83.04          \\
 FSBI (Ours)& \textcolor{black}{ \cmark }         &      \textcolor{red}{ \xmark }        &        \textcolor{red}{ \xmark }         &        \textcolor{red}{ \xmark }         &   \textbf{84.68}          \\ \hline
SBI~\cite{shiohara2022detecting} &  \textcolor{red}{ \xmark }   &  \cmark             &        \textcolor{red}{ \xmark }           &      \textcolor{red}{ \xmark }           & 78.36          \\
FSBI (Ours) & \textcolor{red}{ \xmark }     & \textcolor{black}{ \cmark }            &        \textcolor{red}{ \xmark }         &           \textcolor{red}{ \xmark }      &   \textbf{81.72}          \\ \hline
SBI~\cite{shiohara2022detecting}     &   \textcolor{red}{ \xmark }   &       \textcolor{red}{ \xmark }       & \textcolor{black}{ \cmark }               &      \textcolor{red}{ \xmark }           & 93.10          \\
FSBI (Ours) & \textcolor{red}{ \xmark }        &        \textcolor{red}{ \xmark }      & \textcolor{black}{ \cmark }               &         \textcolor{red}{ \xmark }        & \textbf{94.29}          \\ \hline
SBI~\cite{shiohara2022detecting} & \textcolor{red}{ \xmark }         &   \textcolor{red}{ \xmark }           &       \textcolor{red}{ \xmark }          & \textcolor{black}{ \cmark }               & 94.66          \\ 

FSBI (Ours) & \textcolor{red}{ \xmark }     &       \textcolor{red}{ \xmark }       &         \textcolor{red}{ \xmark }        & \textcolor{black}{ \cmark }               & \textbf{95.49} \\ \bottomrule
\end{tabular}
\end{table*} 

\section{\bfseries Conclusion} \label{sec:conclusion}
In this work, we present a novel deepfake detection method, ESBI, that significantly outperforms existing methodologies. Our approach capitalizes on the synergies of SBI and DWT, enabling our model to discern subtle artifacts in manipulated content. Through extensive experiments and ablation studies, we demonstrate the effectiveness of our proposed method, achieving a remarkable AUC of 95.49\%. The strategic combination of FSBI with the EfficientNetB5 architecture showcases the potential for enhanced generalization and adaptability. Our results set a new benchmark in cross-dataset evaluation, underscoring the efficacy of our approach in combating deepfake threats across diverse datasets.


%



\section*{Acknowledgment}

The authors would like to acknowledge the support received from the Saudi Data and AI Authority (SDAIA) and King Fahd University of Petroleum and Minerals (KFUPM) under the SDAIA-KFUPM Joint Research Center for Artificial Intelligence Grant no. JRC-AI-RFP-14.

\ifCLASSOPTIONcaptionsoff
  \newpage
\fi

\end{document}